\documentclass[runningheads]{llncs}
\usepackage{natbib}%\usepackage[numbers,sort]{natbib}
\usepackage{apalike}
\usepackage{amsmath}
\usepackage{amsfonts}
\usepackage{amssymb}
\newcommand{\uni}{\cup} % set union
\title{Maximizing Expected Impact in an Agent Reputation Network -- Technical Report}  % put your title here!
\author{
Gavin Rens\inst{1} \and
Abhaya Nayak\inst{2} \and
Thomas Meyer\inst{1}
}
\institute{University of Cape Town, South Africa\\
\email{\{grens,tmeyer\}@cs.uct.ac.za}\\
\and
Macquarie University, Sydney, Australia\\
\email{abhaya.nayak@mq.edu.au}
}

\begin{document}

\maketitle

\begin{abstract}
Many multi-agent systems (MASs) are situated in stochastic environments. Some such systems that are based on the partially observable Markov decision process (POMDP) do not take the benevolence of other agents for granted. We propose a new POMDP-based framework which is general enough for the specification of a variety of stochastic MAS domains involving the impact of agents on each other's reputations. A unique feature of this framework is that actions are specified as either undirected (regular) or directed (towards a particular agent), and a new directed transition function is provided for modeling the effects of reputation in interactions. Assuming that an agent must maintain a good enough reputation to survive in the network, a planning algorithm is developed for an agent to select optimal actions in stochastic MASs. Preliminary evaluation is provided via an example specification and by determining the algorithm's complexity.

\end{abstract}

\section{Introduction}

Autonomous (synthetic) agents need to deal with questions of trust and reputation in diverse domains such as e-commerce platforms, crowdsourcing systems, online virtual worlds, and P2P file sharing systems \citep{yslml13,ps13}, wireless sensor networks \citep{mc10}, the semantic web \citep{ag07} and distributed AI/multi-agent systems \citep{ss05b}.
Yet we see very few computational trust/reputation frameworks which can handle uncertainty in actions and observations in a principled way and which are general enough to be useful in several domains. A partially observable Markov decision process (POMDP) \citep{m82,l91} is an abstract mathematical model for reasoning about the utility of sequences of actions in stochastic domains. Although its abstract nature allows it to be applied to various domains where sequential decision-making is required, a POMDP is typically used to model a single agent.
We propose an extension to the POMDP, which has the potential to be quite generally applicable in stochastic multi-agent systems where trust and reputation are an issue. We call the proposed model \textit{Reputation Network POMDP} (\textit{RepNet-POMDP} or simply \textit{RepNet}).

As is done by \cite{psdp10}, we distinguish between the \textit{image} one agent has in the eyes of another and the \textit{reputation} with which an agent is associated, given its aggregated image in the eyes of a group of agents.
The unique features of a RepNet are: (i) it distinguishes between undirected (regular) actions and directed actions (towards a particular agent), (ii) besides the regular state transition function, it has a directed transition function for modeling the effects of reputation in interactions and (iii) its definition (and usability) is arguably more intuitive than similar frameworks. Furthermore, we suggest methods for updating agents' image of each other, for learning action distributions of other agents, and for determining perceived reputations from images. We present a planning algorithm for an agent to select optimal actions in a network where reputation makes a difference.

Many multi-agent systems need a rich framework to represent uncertainty, especially when they are associated with the real world. What makes RepNets different from most other frameworks for multi-agent reasoning about trust and reputation is that uncertainty is treated as a primary concern. Many aspects of the network are modeled as probability distributions, for instance, state transitions, possible observations, possible actions, and possible states. With this richness in representation comes the usual higher complexity in reasoning.

We start by covering the relevant POMDP theory, for background.
This is followed by a section formally defining the proposed framework, including a proposal for how reputation can be computed and learnt (updated) and an algorithm for determining the actions which lead to optimal benefit.
The \textit{optimal impact} algorithm is evaluated in terms of computational complexity. We take a closer look at the RepNet model structure by working out the specification of a simple domain.
Finally, before our concluding remarks, we briefly review the work most related to trust and reputation in groups of agents, especially where there is much uncertainty.

\section{Background - The POMDP}

A POMDP is a tuple $\langle S,$ $A,$ $T,$ $\Omega,$ $O,$ $R,$ $b^0 \rangle$, where
$S$ is a finite set of states,
$A$ is a finite set of actions,
$T$ is a transition function s.t.\ $T(s,a,s')$ is the probability that $a$ executed in $s$ will take an agent to $s'$,
$\Omega$ is a finite set of observations,
$O$ is an observation function s.t.\ $O(a,o,s)$ is the probability that observation $o$ due to action $a$ is perceived in $s$,
$R$ is a reward function s.t.\ $R(s,a,s')$ is the (immediate) reward the agent experiences when it executes action $a$ in state $s$ and ends up in state $s'$, and
$b^0\in \Delta(S)$ is the initial probability distribution over states when the agent is deployed.\footnote{$\Delta(X)$ is the set of probability distributions over elements in $X$.}

In POMDP theory, a probability distribution over states is called a \textit{belief state}.
Formally, a belief state is a function $b:S\to[0,1]$ such that $\sum_{s\in S}b(s)=1$.
The \textit{state estimation} function is used by an agent to update its belief state.
An agent with current belief state $b_\mathit{cur}$ will have new belief state $b_\mathit{new}=SE(a,o,b_\mathit{cur})$ after performing action $a$ and making observation $o$, where
$SE(a,o,b_\mathit{cur}):= \{(s',p)\mid s'\in S\}$
such that
\[
p = \frac{O(a,o,s')\sum_{s\in S}T(s,a,s')b_\mathit{cur}(s)}{P(o\mid a, b_\mathit{cur})}
\]
is the probability of perceiving $o$ in the belief state `reached' after performing $a$ in $b$ ($P(o\mid a, b):= \sum_{s'\in S}O(a,o,s')\sum_{s\in S}T(s,a,s')b(s)$ is a normalizing factor).

The expected reward -- with respect to some belief state $b$ -- an agent gets for executing action $a$ is computed as
$
\sum_{s\in S}  R(a,s)b(s).
$

An agent should maximize the rewards it expects to receive in the future. Rewards farther into the future are conventionally `discounted' because, intuitively, situations in the distant future are less likely to be experienced. Hence, a \textit{discount factor} $\gamma\in[0,1]$ is applied when determining the value of sequences of actions.
The optimal value of belief state $b$ given an horizon of $k$ steps is defined as the optimal belief state value function $V^*$:
\begin{align*}
& V^*(b,k):= \max_{a\in A}\big\{\sum_{s\in S}R(a,s)b(s)+ \gamma\sum_{o\in \Omega}P(o\mid a,b)V^*(SE(a,o,b),k-1)\big\},\\
& V^*(b,1):= \max_{a\in A}\big\{\sum_{s\in S}R(a,s)b(s)\big\}.
\end{align*}

\section{RepNet-POMDP - a Proposal}

We shall first introduce the basic structure of a RepNet-POMDP, then discuss matters relating to image and reputation, followed by our proposal of two instantiation of the image update function, then explain how action distributions can be learnt and finally, develop a definition (and thus an algorithm) for computing optimal behaviour in RepNets.

\subsection{The Basis}

The components of the RepNet structure will first be introduced briefly, followed by a detailed discussion of each component. A RepNet-POMDP is defined as a pair of tuples $\langle \mathit{System}, \mathit{Agents} \rangle$.
$\mathit{System}$ specifies the aspects of the network that apply to all agents; global knowledge shared by all agents.
\[\mathit{System} := \langle G,S,A,\Omega,I,U \rangle,\]
where
\begin{itemize}
\item $G$ is a finite set of agents $\{g, h, i, \ldots\}$.
\item $S$ is a finite set of states.
\item $A$ is the union of finite disjoint sets of \textit{directed} actions $A^d$ and \textit{undirected} actions $A^u$.
\item $\Omega$ is a finite set of observations.
\item $I: G\times S\times G\times S \times A\to [-1,1]$ is an \textit{impact} function s.t.\ $I(g,s,h,s',a)$ is the impact on $g$ in $s$ due to $h$ in $s'$ performing action $a$.
\item $U:[0,1]\times [-1,1]\times[-1,1]\to[-1,1]$ is an \textit{image update} function used by agents when updating their image profiles s.t.\ $U(\alpha,r,i)$ is the new image level given learning rate $\alpha$, current image level $r$ and current impact $i$.
\end{itemize}

$\mathit{Agents}$ specifies the names and subjective knowledge of the individual agents; individual identifiers and beliefs per agent.
\[\mathit{Agents} := \langle \{T_g\},\{DT_g\},\{O_g\},\{AD_g^0\},\{Img_g^0\},\{B_g^0\} \rangle,\footnote{Notation $\{X_g\}$ is shorthand for $\{X_g\mid g\in G\}$. That is, there is a function $X$ for each agent in $G$.}\]
where
\begin{itemize}
\item $T_g:S\times A^u\times S\to[0,1]$ is the \textit{transition} function of agent $g$.
\item $DT_g:S\times A^d\times [-1,1]\times S\to[0,1]$ is the \textit{directed transition} function of agent $g$ s.t. $DT_g(s,a,r,s')$ is the probability that agent $g$ executing an action $a$ in state $s$ (directed towards some agent $h$) will take $g$ to state $s'$, while $g$ believes that agent $h$ perceives $g$'s reputation to be at level $r$. $DT_g(s,a,r,s')=P(s'\mid g,s,a,r)$, hence $\sum_{s'\in S}DT_g(s,a,r,s')$ $=$ $1$, given some current state $s$, some reputation level $r$ and some directed action $a$ of $g$.
\item $O_g$ is $g$'s observation function s.t.\ $O_g(a,o,s)$ is the probability that observation $o$ due to action $a$ is perceived by $g$ in $s$.
\item $AD_g^0:G\times S\to \Delta(A)$ is agent $g$'s initial \textit{action distribution} providing $g$ with a probability distribution over actions for each agent in each state.
\item $Img_g^0: G\times G \to [-1,1]$ is $g$'s initial image profile. $Img_g(h,i)$ is agent $h$'s image of agent $i$, according to $g$.
\item $B_g^0:G\to\Delta(S)$ is $g$'s initial mapping from agents to belief states.
\end{itemize}

The agents in $G$ are thought of as forming a linked group who can influence each other positively or negatively and who cannot be influenced by agents outside the network. % This is obviously an idealized setting, but might be useful for several applications.
It is assumed that all action execution is synchronous, that is, one agent executes one action if and only if all agents execute one action. All actions are assumed to have an equal duration and to finish before the next actions are executed. The immediate effects of actions are also assumed to have occurred before the next actions. That is, for every agent's action, the successor state is reached, reputation updated and impact completed before the next round of actions is executed. We shall call these rounds of actions \textit{steps}.

All agents share knowledge of who the agents in the network are, what the set of possible states is ($S$), what actions can possibly be performed ($A$), impact of actions ($I$), image update function ($U$), the set of possible observations ($\Omega$) and the likelihoods of perceiving them in various conditions. All other components of the structure relate to individual agents and how they model some aspect of the network: dynamics of their actions ($T_g$ and $DT_g$) and observations ($O_g$), likelihood of actions of other agents ($AD_g$), beliefs about reputation ($Img_g$) and their initial belief states ($B_g$).

In this formalism, only the action distributions ($AD_g$), image profiles ($Img_g$) and set of belief states ($B_g$) change. All other models remain fixed.

An agent should maintain an image profile for all other agents in the network in order to guide its own behaviour. An image profile is an assignment of image levels between every ordered pair of agents. For instance, if (according to $g$) $h$'s image of $i$ ($Img_g(i,h)$) is, on average, low, $g$ should avoid interactions with $i$ if $g$ has a good image of $h$ ($Img_g(h,g)$).
Note that agents' multi-lateral image is not common knowledge in the network. Hence, each agent has only an \textit{opinion} about each pair of agent's image as deemed by each other agent. In general, it is not necessary that $Img_g(g,h)=Img_h(g,h)$. Nor is it necessary that $Img_g(h,i)=Img_g(i,h)$.
% Given that the proposed formalism has a central concept (belief/opinion) of \textit{reputation}, $DT$ is conditioned on reputation levels. 

$Img_g(h,i)$ changes as agent $g$ learns how agent $i$ `treats' its network neighbour $h$. Agent $g$ uses $U$ to manage the update of its levels of reputation as deemed by other agents.
An agent needs to have a strategy how to build up its image profile of each other agent. How should an agent adjust its opinion of another agent if it keeps on being impressed by that agent? Formally, there is a maximum image level of 1. Should a well-behaved agent get an image of 1 after relatively few actions and then have its image stay at 1? Or should such a `good' agent's image gradually approach 1? If so, how gradually? We decided to define the image update function $U$ common to all agents for the sake of simplicity, while introducing the RepNet-POMDP framework. This assumption can easily be relaxed by defining $U$ differently for each agent.

Actually, we define directed transitions to be conditioned on reputation (derived from images):
Suppose $g$ wants to trade with $h$. Agent $g$ could perform a $\mathtt{tradeWith}{\_}h$ action. But if $h$ deems $g$'s reputation to be low, $h$ would not want to trade with $g$. This is an example where the effect of an action by one agent ($g$) depends on its level of reputation as perceived by the recipient of the action ($h$). Note that it does not make sense to condition the transition probability on the reputation level of the recipient as perceived by the actor ($h$'s reputation as perceived by $g$ in this example): The effect of an action by $g$ should have nothing to do with $h$'s image levels, given the \textit{action is already committed to} by $g$. However, the effect of an action committed to (especially one directed towards a particular agent) may well depend on the actor's ($g$'s) reputation levels; $h$ may react (effect of the action) differently depending on $g$'s reputation.

Continuing with the example, assume $s'$ is a state in which $g$ gets what it wanted out of a trade with $h$, and $s$ is a state in which $g$ is ready to trade. Then $DT(s,\mathtt{tradeWith}{\_}h,-0.6,s')$ might equal 0.1 due to $h$'s inferred unwillingness to trade with $g$ due to $g$'s current bad reputation ($-0.6$) as deemed by $h$. On the other hand, $DT(s,\mathtt{tradeWith}{\_}h,0.6,s')$ might equal 0.9 due to $g$'s high esteem ($0.6$) as deemed by $g$ and thus inferred willingness to trade with $g$.

It makes sense to talk about the impact/effect of an action ($a$) from one state ($s$) to another ($s'$): $I(g,s,h,s',a)$. Suppose you (agent $g$) are in state $s$ where you have a high energy level, and I (agent $h$) am in state $s'$ where I have a low energy level. If you push me (action $a$), you can push harder than usual and I can recover or defend myself less than usual. The impact of the action is thus dependent on both the state in which the action is executed and the state being impinged upon.
% The arguments g and h could not be left out because ...

It is assumed that every agent $g$ has some (probabilistic) idea about what actions its neighbours will perform in a given state. As stated in the definition of the RepNet-POMDP structure, $AD_g(h,s)$ is a distribution over the actions in $A$ that $h$ could take when in state $s$. It is debatable how realistic it would be to have a single action model $AD(h,s)$ common to all agents. Every agent $g$ perceives different signals, depending on their state. Every agent would thus learn a different action distribution for its neighbours. One method for learning these distributions is proposed later in this section.

The other component of the structure which changes is $B_g$; every agent ($g$) maintains a probability distribution over states for every agent in $G$ (including itself). That is, for every agent $g$, its belief state for every agent $h$ ($B_g(h)$) is maintained and updated. In other words, every agent maintains a belief state representing `where' it thinks the other agents (incl. itself) are. As actions are performed, every $g$ updates these distributions of itself and its neighbours. In POMDP theory, probability distributions over states are called \textit{belief states}. $B_g$ changes via `normal' state estimation as in regular POMDP theory. The state estimation (update) theory for RepNet-POMDPs is explained in the next subsection.

%It is important to note that not every agent mentioned in $G$ needs to follow an optimal policy, that is, some agents may act unpredictably. Moreover, some agents' transition and observation functions ... ???

\subsection{Image and Reputation in RepNets}

There are many ways in which an agent can compute reputations, given the components of a RepNet-POMDP. In this section, we investigate one approach.

The notions of trust and reputation are typically considered to be different \citep{ag07,sp09,mc10}: reputation informs trust with reputation itself being an agregation of indirect information, namely, evalution of an agent which has propagated through a network. However, it has also been argued that trust and reputation are not easily distinguished \citep[Sect. 7.1, e.g.]{psdp10}. At a basic level, \textit{image} is an opinion of another agent which is built up from direct experience with that agent \citep{cp02,psdp10}. Direct experience or direct interaction with an agent is sometimes viewed as a kind of reputation, however, we feel it is useful to distinguish image as a personal experience, reputation as propagated image valuations, and trust as reputation plus other sources of information concerning an agent's willingness to be vulnerable when dealing with the trustee.

Recall that $AD_g(h,s)$ is the probability distribution over actions $g$ believes $h$ executes in $s$. In other words, $AD_g(h,s)(a)$ is the probability of $a$ being executed by $h$ in $s$ according to $g$.
The image that an agent $i$ in state $s^i$ has of an agent $h$ in state $s^h$ can be modeled as
\[
\sum_{a\in A}\big[\delta AD_g(i,s^i)(a)I(h,s^h,i,s^i,a)+ (1-\delta)AD_g(h,s^h)(a)I(i,s^i,h,s^h,a)\big],
\]
where $\delta\in[0,1]$ trades off the importance of the impacts on $h$ and impacts due to $h$. As $\delta$ tends to 1, more importance is given to how $i$ treats $h$ and less to how $h$ treats $i$. Recall that $B_g$ is the set of current belief states of all agents in the network, according to $g$. Hence, $B_g(i)$ is a belief state, and $B_g(i)(s)$ is the probability of $i$ being in $s$, according to $g$. For better readability, we might denote $B_g(i)$ as $b^g_i$.
Agent $g$ perceives at some instant that $i$'s image of $h$ is
\begin{align}
Image_g(h,i,B_g):= & \sum_{s^h\in S}b^g_h(s^h)\sum_{s^i\in S}b^g_i(s^i)\sum_{a\in A}\big[\delta AD_g(i,s^i)(a)I(h,s^h,i,s^i,a)\nonumber \\
& + (1-\delta)AD_g(h,s^h)(a)I(i,s^i,h,s^h,a)\big]\label{eq:image}.
\end{align}
In \eqref{eq:image}, the uncertainty of agents $h$ and $i$'s states are taken into account.
Note that this perceived image is independent of $g$'s state.

%Finally, expanding \eqref{eq:inst-trustin-uncertain-expected}, $g$'s instantaneous perceived reputation in $h$ is defined as
%\[
%PTI(g,h,B_g):=\sum_{s^h\in S}b^g_h(s^h)\sum_{a\in A}\sum_{i\in G}\sum_{s^i\in S}b^g_i(s^i)AD_g(h,s^h)(a)I_h(s^h,a,s^i).
%\]
Now we can define the new image of one agent $h$ with respect to another $i$, as perceived by a particular agent $g$:
\[
U(\alpha,Img_g(h,i),Image_g(h,i,B_g)),
\]
where $\alpha$ is a learning rate supplied from outside.
Just as $SE$ updates an agent's belief state, the \textit{image expectation} function $IE(g,Img_g,\alpha,B_g) := Img'_g$ updates an agent's image profile. That is, given $g$'s set of belief states $B_g$, for all $h,i\in G$,
\[
Img'_g(h,i) = U(\alpha,Img_g(h,i),Image_g(h,i,B_g)).
\]

An agent $g$ could form its opinion about $h$ in at least three ways: (1) by observing how other agents treat $h$, (2) by observing how $h$ treats other agents and (3) by noting other agents' opinion of $h$. But $g$ must also consider the reasons for actions and opinions: Agent $i$ might perform an action with a negative impact on $h$ because $i$ believes $h$ has a bad reputation or simply because $i$ is bad. We define reputation as
\[
\mathit{RepOf}_g(h):= \frac{1}{|G|}\big[Img_g(h,g) + \sum_{i\in G, i\neq g}Img_g(h,i)\times Img_g(i,g)\big].
\]
Here, we have assumed that it does not make sense to weight $Img_g(h,g)$ by $Img_g(g,g)$ because it makes no sense to weight one's opinion about $h$'s image by one's opinion of one's own image. Hence, $Img_g(h,g)$ is implicitly weighted by 1.

To get a better feel for the behaviour of $\mathit{RepOf}_g(h)$, consider the following table.
\begin{table}[h!]
\centering
\begin{tabular}{|c|c|c|}
\hline
$Img_g(h,i)$ & $Img_g(i,g)$ & $Img_g(h,i)\times Img_g(i,g)$\\
\hline
-0.5 & -0.5 & 0.25\\
0.0 & -0.5 &0.0 \\
0.5 & -0.5 & -0.25\\
-0.5 & 0.0 & 0.0\\
0.0 & 0.0 & 0.0\\
0.5 & 0.0 & 0.0\\
-0.5 & 0.5 & -0.25\\
0.0 & 0.5 & 0.0\\
0.5 & 0.5 & 0.25\\
\hline
\end{tabular}
\end{table}
The third column can be thought of as the real reputation of $h$ according to what $g$ believes $i$ thinks of $h$. Notice that if $i$ has a negative image level, then $g$ regards $h$ in an opposite way to how $i$ regards $h$. That is, an agent with a negative image is regarded as purposefully misleading. Moreover, the degree of negativity of $i$'s image gives the degree to which $g$ regards $h$ in the opposite light. If $h$ has zero image according to $i$, then $g$ cannot assign a reputation to $h$, no matter what $g$ thinks of $i$. Similarly, if $i$ has zero image, then $g$ cannot assign any reputation to $h$, no matter what $i$ thinks of $h$. When $i$ has an image close to zero (according to $g$), it is like assigning very little informational value to what $i$ says. And if $i$ has high image (close to 1), then $i$'s opinions are highly trusted, but if $i$ has low image (close to -1), then $i$'s opinions are highly distrusted, informing $g$ to believe the complete opposite of what $i$ proclaims.

The question is, Where did $g$ get $i$'s image? What if, for instance, $g$ hears from trusted agent $j$ that $i$'s reputation/image is opposite to what $g$ thought of $i$'s reputation? There are several ways to remedy this uncertainty. One way is to iteratively seek and weight the reputation of agents to some `depth'. This method has technical difficulties which we do not try to address now. The simple approach above partly solves the problem in two ways. (1) $i$'s reputation is only one of all the reputations considered by $g$, and $g$ takes the \textit{average} of all agents' opinions of $g$ to come to a conclusion of what to think of $h$ ($h$'s reputation according to $g$). (2) Reputation is also informed by actual activity, as perceived by each agent $g$. Hence, every agent builds up more accurate opinions of other agents, according to their \textit{activities} (not only what others say about others). Activities inform image and image informs reputation.

Suppose a step has just occurred and $g$ perceives $o$. Let $b^g_{h}$ be the belief state $g$ assigns to $h$.
The new belief state of some agent $g$ according to itself is defined by
\begin{equation}
\label{def:ose}
OSE(a,o,(b^g_g)_\mathit{cur}):= \{(s',p)\mid s'\in S\}
\end{equation}
such that
\[
p = \frac{O(a,o,s')\sum_{s\in S}T^\mathit{du}_g(s,a,s',g)(b^g_h)_\mathit{cur}(s)}{P(o\mid a, (b^g_g)_\mathit{cur})}
\]
and, recalling that $A^d$ and $A^u$ are the sets of directed, respectively, undirected actions,
\[
T^\mathit{du}_g(s,a,s',h):= \left\lbrace
\begin{array}{cl}
DT(s,a,\mathit{RepOf}_g(h),s')&\mbox{if }a\in A^d\\
T(s,a,s')&\mbox{if }a\in A^u
\end{array}
\right.
\]
and $P(o\mid a, b)$ is a normalizing constant.
We call definition~\eqref{def:ose} the \emph{objective state estimation} function (of agent $g$).

The new belief state of $h$ ($\neq g$) according to $g$ is defined by
\begin{equation}
\label{def:sse}
SSE(g,h,o,(b^g_h)_\mathit{cur}):= \{(s',p)\mid s'\in S\}
\end{equation}
such that
\[
p = \frac{\sum_{a\in A}O(a,o,s')\sum_{s\in S}T^\mathit{du}_g(s,a,s',h)(b^g_h)_\mathit{cur}(s)AD_g(h,s)(a)}{P(o\mid AD_g, (b^g_h)_\mathit{cur})}
\]
and $P(o\mid AD_g, b)$ is a normalizing constant.
We call definition~\eqref{def:sse} the \emph{subjective state estimation} function (of agent $h$ other than $g$).

An agent $g$ uses \eqref{def:sse} when updating its belief about other agents' belief states, else $g$ uses \eqref{def:ose} to update its own belief state.
Let $BSE(g,a,o,B_g)$ be the set of belief states of all agents (from $g$'s perspective) after the next step, determined from the current set of belief states $B_g$, given agent $g$ executed $a$ and perceived $o$. That is, $BSE(g,a,o,B_g)$ is defined as
%\begin{align*}
%&BSE(g,a,o,B_g):= \{(h,b^g_h)\mid h\in G,h\neq g,\\
%&\qquad b^g_h = SSE(g,h,o,B_g(h))\} \uni \{(g,OSE(a,o,B_g(g)))\}.
%\end{align*}
\[\{(h,b^g_h)\mid h\in G,h\neq g,b^g_h = SSE(g,h,o,B_g(h))\} \uni \{(g,OSE(a,o,B_g(g)))\}.\]

\subsection{Possible Instantiations of $U$}

One instantiation implements the idea that if the impact is positive, a fraction of the reputation required to reach full reputation (1) is added to the current reputation level, and if the impact is negative, a fraction of the reputation required to reach full distrust (-1) is subtracted from the current reputation level. The fraction added or subtracted is proportional to the learning rate $\alpha$ and the magnitude of the impact ($|i|$):
\[
U(\alpha,r,i) := \left\lbrace
\begin{array}{rl}
r + \alpha(1-r)i & \mbox{if } i\geq 0\\
r + \alpha(r+1)i & \mbox{if } i< 0
\end{array}
\right.
\]
We shall refer to this instantiation of $U$ as \textit{difference update}.
Suppose agent $g$ has no opinion about the agent causing the impact (i.e., $r=0$). And suppose $\alpha$ is 0.5. Then
\begin{itemize}
\item if $i=1$, $U(0.5,0,1)=0.5$
\item if $i=0.5$, $U(0.5,0,0.5)=0.25$
\item if $i=-0.5$, $U(0.5,0,-0.5)=-0.25$
\item if $i=-1$, $U(0.5,0,-1)=-0.5$.
\end{itemize}

Another instantiation implements the idea that if the impact is positive, reputation increases until full reputation (1) is reached, and if the impact is negative, reputation is decreased until full distrust (-1) is reached:
\[
U(\alpha,r,i) := \left\lbrace
\begin{array}{rl}
1 & \mbox{if } r+\alpha i>1\\
-1 & \mbox{if } r+\alpha i<-1\\
r+\alpha i & \mbox{otherwise.}
\end{array}
\right.
\]
We shall refer to this instantiation of $U$ as \textit{saturation update}.

Suppose $\alpha$ is 0.5. Suppose $i=1$ for fifty steps (due to a particular agent $h$'s actions). After these fifty steps, $g$'s reputation in $h$ will be 1. Then suppose that for steps 51 and 52, $i=-1$. Then $g$'s reputation in $h$ will be 0 (neutral). However, if $i$ were 1 for only four steps (instead of fifty), and then $-1$ for two steps, then $g$'s reputation in $h$ would still be 0. This might seem counter-intuitive: In the former case, $h$ seems to have earned more reputation than in the latter case, although $g$ ends up having a neutral opinion about $h$ in both cases. On the other hand, one could argue that once a person (agent) is fully trusted, their future positive actions do not add much to one's opinion of the person. However, as soon as that person does one or two things deemed negative, one's reputation in that person rapidly decreases.
High (and low) reputation levels reached via difference update need to be `earned', and these levels are less sensitive to occasional `out-of-character' actions than levels maintained via saturation update.

There are of course many more possible instantiations of $U$. Just as the reward function in POMDPs needs to be tailored for the domain, $U$ needs to be tailored for the RepNet-POMDP domain. Choosing between difference, saturation or some other update function will be domain dependent.

\subsection{Learning Action Distributions}

In this section, we present one way in which an agent can incrementally learn what actions other agents are likely to perform. That is, we propose how some agent $g$ can learn the probability of some agent $h$ performing some action in some state. The agent learns by Bayesian conditionalization on observations.

We use the update (learning) rule
\[
\forall a\in A, AD_g(h,s)(a) \gets P_g(a\mid o,h,s),
\]
where, according Bayes Rule,
\begin{eqnarray*}
P_g(a\mid o,h,s) &=& \frac{P_g(o\mid a,h,s)P_g(a\mid h,s)}{P_g(o\mid h,s)}\\
&=& \frac{P_g(o\mid a,h,s)P_g(a\mid h,s)}{\sum_{a'\in A}P_g(o\mid a',h,s)P_g(a'\mid h,s)}.
\end{eqnarray*}

We would like to replace $P_g(o\mid a,h,s)$ with $O(a,o,s)$, but we may not do a direct replacement: $a$ is assumed performed in $s$, but $o$ is perceived in the state reached via $a$ from $s$. Therefore, $P_g(o\mid a,h,s)$ is approximated by
\[
\sum_{s'\in S}T^\mathit{du}(s,a,s')O(a,o,s').
\]
$P_g(a\mid h,s)$ is replaced with $AD_g(h,s)(a)$. Hence, we define the \textit{action distribution expectation} function as $ADE(g,o,AD_g)$ $:=$ $AD'_g$ such that, for all $h\in G$, $s\in S$ and $a\in A$,
\begin{align*}
&AD'_g(h,s)(a) =\\
&\qquad\frac{\sum_{s'\in S}T^\mathit{du}(s,a,s')O(a,o,s')AD_g(h,s)(a)}{\sum_{a'\in A}\sum_{s'\in S}T^\mathit{du}(s,a',s')O(a',o,s')AD_g(h,s)(a')}.
\end{align*}

\subsection{Optimal Behaviour in RepNets}

Advancement of an agent in RepNet-POMDPs is measured by the total impact on the agent. An agent might want to maximize the network's (positive) impact on it after several steps in the system.

An agent $g$ currently in $s^g$ can predict agent $h$'s action $a$ performed in $s^h$ to cause an instantaneous impact on it ($g$).
The \textit{expected} instantaneous impact of $h$ in $s^h$ on $g$ in $s^g$ is calculated as
\[
\sum_{a\in A}AD_g(h,s^h)(a)I(g,s^g,h,s^h,a).
\]
We abbreviate the un-normalized perceived impact on $g$ in $s^g$ by the network neighbours
\[
\sum_{h\in G,h\neq g}\sum_{s^h\in S}B_g(h)(s^h)\sum_{a\in A}I(g,s^g,h,s^h,a)AD_g(h,s^h)(a)
\] as $PIN(g,s^g,AD_g,B_g)$.
That is, $PIN(g,s^g,B_g)$ is the expected impact other agents have on $g$ in a particular state, with respect to the neighbours' possible actions and their belief states.

The self impact on $g$ executing $a$ in $s^g$ is defined as $I(g,s^g,g,s^g,a)$. Here we take it that every agent knows which action it performs.
We thus define the total perceived impact on $g$ by the network as
\[
PI_\mathit{tot}(g,a,B_g):=\frac{1}{|G|}\sum_{s^g\in S}B_g(g)(s^g)\big(PIN(g,s^g,AD_g,B_g) + I(g,s^g,g,s^g,a)\big).
\]

Intuitively, an agent $g$ can choose its next action so as to maximize the total impact all agents will have on it in the future.
Then the optimal impact function w.r.t. $g$ over the next $k$ steps is defined as
\begin{align*}
OI(g,AD_g,Img_g,B_g,k):= & \max_{a\in A}\big\{PI_\mathit{tot}(g,a,B_g)\\
& + \gamma\sum_{o\in \Omega}P(o\mid a,B_g)OI(g,AD'_g,Img'_g,B'_g,k-1)\big\},\\
OI(g,AD_g,Img_g,B_g,1):= & \max_{a\in A}\big\{PI_\mathit{tot}(g,a,B_g)\big\},
\end{align*}
where $AD'_g$ is $ADE(g,o,AD_g)$, $Img'_g$ is $IE(g,Img_g,\alpha,B_g)$ and $B'_g$ is $BSE(g,a,o,B_g)$.
%The definition above has a very similar form to that of the optimal value function $V^*$ of (regular) POMDP theory.

\subsection{Complexity of Computing Optimal Behaviour}

Every agent in a network is assumed to have its own computing resources.
For an individual agent, the number of computations involved in determining\\ $OI(g,AD_g,Img_g,B_g,k)$ is approximately
\begin{equation}
\label{eq:compcompOI}
(PI_\mathit{tot} + P_\mathit{SSE} + ADE + IE + BSE)\sum^{k-1}_{\ell=1}(|A||\Omega|)^\ell,
\end{equation}
where
\begin{itemize}
\item $PI_\mathit{tot}$ denotes $O(|G||S|^2|A|)$,
\item $P_\mathit{SSE}$ denotes the complexity of $P(o\mid a,B_g)$ which is $O(|A||S|^2)$,
\item $ADE$ denotes $O(|G||S|^2|A|^2)$,
\item $IE$ denotes $O(|G|^2|S|^2|A|)$,
\item $BSE$ denotes $O(2|G||S|^2)$.
\end{itemize}
By the identity of geometric series, \eqref{eq:compcompOI} equals
\begin{align*}
&(PI_\mathit{tot} + P_\mathit{SSE} + ADE + IE + BSE)\frac{|A||\Omega|(1-(|A||\Omega|)^{k-1})}{1-|A||\Omega|}\\
&\quad= O((|G||S|^2|A|+|A||S|^2+|G||S|^2|A|^2+|G|^2|S|^2|A|+2|G||S|^2)|A|^k|\Omega|^k)\\
&\quad= O((|G||S|^2|A|^2+|G|^2|S|^2|A|)|A|^k|\Omega|^k)\\
&\quad= O(|G||S|^2|A|(|A|+|G|)|A|^k|\Omega|^k)\\
&\quad= O(|A|^{k+1}|\Omega|^k|S|^2|G|(|A|+|G|)).
\end{align*}

Compared to the $O(|A|^k|\Omega|^k|S|^2)$ for $V^*$ for regular finite horizon planning, we see that computing the optimal impact (for an individual agent) in a RepNet-POMDP is in the order of $|A|(|A| + |G|)$ times more complex. If, however, computation is centralized on one CPU, and assuming that every agent requires planning (i.e., every agent is an AI), then the complexity of finite horizon planning in a RepNet-POMDP is $O(|A|^{k+1}|\Omega|^k|S|^2|G|^2(|A|+|G|)).$

\section{Related Work}

We separate this section into three groups: Section~\ref{sec:i}: frameworks that include notions of quantitative uncertainty, but that do not involve the POMDP model, Section~\ref{sec:ii}: frameworks that employ POMDPs as a component, but cannot be specified as a POMDP or variant thereof alone, and Section~\ref{sec:iii}: frameworks that are variants of the POMDP (and thus most related to the present work). In each subsection, we provide a short review of related work and then compare it to the RepNet framework.

\subsection{Non-POMDP frameworks that include notions of quantitative uncertainty}\label{sec:i}

We briefly review two papers under this category.

\cite{ys02} develop an (uncertain) evidential model of reputation management based on the Dempster-Shafer theory.
%These uncertainties are modeled and computed via Dempster-Shafer theory. 
If an agent $g$ has had sufficient interactions with agent $h$, $g$ bases its opinions on the quality of services received during those interactions. If $g$ has had insufficient interactions with $h$, it will seek referrals from (trusted) neighbours, who may be witnesses of interaction with $h$. If the neighbours are not witnesses, they will request witnesses from their neighbours, and so on.
A limitation of this approach is that it models only the uncertainty in the services received and in the trustworthiness of neighbours who provide referrals. It does not model dynamical systems, nor does it allow for stochastic actions and observations.

\cite{psdp10} propose an integration of a cognitive reputation model, called \textit{Repage}, into a BDI agent. They define a many-sorted, first-order, probabilistic logic capable of capturing the semantics of Repage information. The logic includes predicates for belief, image and reputation.
% The belief logic integrates the information coming from Repage in terms if image and reputation, and combines them, defining a typology of agents depending of such combination. 
The logic is used to build a multi-context BDI system where beliefs, desires, intentions and plans interact among each other to perform reasoning.
The authors base their architecture on the social evaluation theory of \citep{cp02} which describes a typology of possible decisions that autonomous agents can make: \textit{Epistemic decisions} cover the dynamics of beliefs regarding image and reputation, that is, decisions about updating and generating evaluations by social groups. \textit{Pragmatic}–\textit{strategic decisions} are are concerned with how to behave with potential partners using social group evaluation information, and thus, how agents use these decisions to reason. \textit{Memetic decisions} refer to the decisions of how and when to spread social evaluations. \cite{psdp10} focus on pragmatic-strategic decisions.
Their logic uses the notion of \textit{roles} for specifying capabilities or services, for instance, quality of product and delivery time. The probabilities of action outcomes and the probabilities of agent roles can be specified.
Probabilistic distributions over image and reputation (for the same agent and role) are combined to generate beliefs that an agent acts on.

With their logic, \cite{psdp10} can specify capabilities or services that our framework cannot. On the other hand, their Repage + BDI architecture cannot model noisy observations or uncertainty in state (belief states).
Whereas the approach of \cite{ys02} keeps a history of a fixed number of interactions to determine ``local belief'' and fixed depth ``referral chains'', our framework is more flexible. Admittedly, our framework may require a learning rate for maintaining `local belief' (which we call \textit{image}).

\subsection{Frameworks that employ POMDPs as a component}\label{sec:ii}

We briefly review two frameworks with a POMDP as component.

\cite{rcp05} aim to construct a principled framework, called \textit{Advisor-POMDP}, for buyers to choose the best seller based on some measure of reputation in a market consisting of autonomous agents: ``A selection of approaches to representing reputation using Dempster-Shafter Theory and Bayesian probability are surveyed and a model for collecting and using reputation is developed using a Partially Observable Markov Decision Process'' \citep{rcp05}.
It is noted that Advisor-POMDP models a single buyer agent, maintaining a vector of reputations of sellers. The buyer may ask for advice from advisors about a seller, who respond with (i.e., the buyer perceives) $\langle rep_i, cf_i\rangle$ where $rep_i$ is the reputation and the certainty factor $cf_i$ is a measure of the epistemic uncertainty of seller $i$. RepNets do not do not utilize such confidence levels.

SALE POMDP \citep{ioz14} is an extension of Advisor-POMDP: It can deal with the seller selection problem by reasoning about advisor quality and/or trustworthiness and selectively querying for information to finally selects a seller with high quality. Moreover, SALE POMDP has a factored formulation which ``allows it to scale to reasonably large seller selection problems without loss in quality'' of agent behaviour \citep{ioz14}.

The major difference between Advisor- and SALE POMDP on the one hand, and RepNets on the other, is that a RepNet has a model for every agent in the network and every agent has a (subjective) view on every other agent's belief state and action likelihood. Advisor- and SALE POMDP do not allow for the modeling of common information shared by all agents in the network, and hence cannot cater for agents having views of other agents' belief state and action likelihood.

\subsection{Frameworks that are based on POMDPs}\label{sec:iii}

Whereas Decentralized POMDPs (DEC-POMDPs) \citep{bzi00} are concerned more with effective collaboration in noisy environment than with self-advancement in a network of potentially unfriendly strangers.
Although each agent has its own actions and observations in a DEC-POMDP, the effects of their \textit{combined} actions and observations are modeled. \cite{acguk13} discuss four notable sublasses of the DEC-POMDP.

Interactive POMDPs (I-POMDPs) \citep{gd05} are for specifying and reasoning about multiple agents, where willingness to cooperate is not assumed. Whereas DEC-POMDP agents do not have a model for every other agent's belief state and action likelihood, I-POMDP agents maintain a model of every other agent (incl. their own). A model that agent $g$ has of agent $h$ is not necessarily the model $h$ has of itself. In other words, agents have beliefs about what other agents believe. The models of other agents also contain models of other agents, recursively. The recursion could theoretically be infinite, but for practical purposes, models are finitely nested.
An agent's state is composed of two parts: a part representing the physical environment and a part representing the (nested) models of other agents. Every agent has its own actions and observations, and as in DEC-POMDPs, every agent has its own transition function for combined actions, and an observation function for combined actions and own observations. Only the physical part of states are involved in transition and observation functions. An agent's reward function models its preferences over physical states \textit{and} models of other agents.
In contract to RepNet agents which keep only the belief state of other agents and action likelihood (not the full model; not recursive), I-POMDPs have an arbitrary model nesting depth. The benefit of deeper nesting is the deeper insights and thus better informed decisions that can be made. However the drawback of deeper nesting is computational complexity (see \cite{dg09} for Monte-Carlo sampling methods for approximate solutions to I-POMDPs). The fact that a RepNet keeps track of other agents' beliefs outside of the set of states $S$ is a form of factorization, which reduces the curse of dimensionality. That is, because other agents' beliefs are factored out of the set of possible states $S$, the cardinality of $S$ is orders of magnitude less.

I-POMDPs and DEC-POMDPs do not have a notion for trust, reputation or image.
Seymour and Peterson \cite{sp09} introduce notions of trust to the I-POMDP, which they call trust-based I-POMDP (TI-POMDP).
A trust model $\tau_i$ (similar to our $Img_g$) is maintained as part of agent states.
They define transition functions from environment states to environment states (via composite actions), which reportedly includes changes to $\tau_i$. 
However, there are several inconsistencies in the presentation of their framework; it is thus hard to compare RepNets to TI-POMDPs. Nonetheless, RepNets are arguably more understandable model structures than I-POMDP and TI-POMDPs, and as with I-POMDPs, the factored formulation of RepNets reduces the curse of dimensionality.

\section{Summary and Future Work}

This paper presented a new framework, called RepNet-POMDP, for agents in a network of self-interested agents to make considered decisions. The framework deals with several kinds of uncertainty and facilitates agents in determining the reputation of other agents. An algorithm was provided for an agent to look ahead several steps in order to choose actions in a way that will influence its reputation so as to maximize the network's positive impact on the agent. We aimed to make the framework easily understandable and generally applicable in systems of multiple, self-interested agents where partial observability and stochasticity of actions are problems.
To illustrate the usefulness of RepNet-POMDPs, the full version of this paper has an example where we set up a trading domain for a network of four agents.

There are of course many more possible instantiations of $U$. Just as the reward function in POMDPs needs to be tailored for the domain, $U$ needs to be tailored for the RepNet-POMDP domain. Choosing between difference, saturation or some other update function will be domain dependent. The full version of this paper provides a detailed discussion.

In the full version of this paper, we present one way in which an agent can incrementally learn what actions other agents are likely to perform. That is, we propose how some agent $g$ can learn the probability of some agent $h$ performing some action in some state. The agent learns by Bayesian conditioning on observations.

There are many ways to define image and reputation. The framework could also have been made more sophisticated, requiring a trust layer. These considerations are left for future work.
We did not discuss how one could, in the RepNet framework, distinguish between a new agent whose image or reputation is unknown and a well-known agent with an average image ($\mathit{RepOf}_g(h)\approx 0$). This is not a failing of the RepNet framework, but discuss it is beyond the scope of this paper.
We have assumed that $\mathit{RepOf}_g$ is the weighted average of all image opinions, but it might be more reasonable for $g$ to take others' opinions less seriously as it builds confidence about images of agents over time via actual activities perceived.
%
%An advantage of the proposed approach for an agent to deciding (via the directed transition function) when to interact with another agent is that it does not rely on third party testimony; third-party trust is thus not an issue. The disadvantage of the approach is that third-party testimony cannot be used in the early stages of an agent's experience in the network system. Yes, an agent $g$ can compute the reputation of another agent, but this still relies on $g$'s \textit{beliefs} about what other agents images.
%
Another aspect of RepNet that needs improvement is reputation (trust) in \textit{context}. The present framework assumes a global context, wherein the learnt reputation of an agent is applicable at all times / in all states. This approach is appropriate only in very narrow or specific domains where there are few degrees of freedom in context. Typically though, context will change, and applicability of reputation of agents will change with context. See, for instance, the work of \cite{n12} and \cite{psdp10}.

Asking for trust/reputation opinions from trusted agents is not possible in RepNet-POMDPs. All reputation info comes through observation; none through direct communication. If the image profiles were part of the states (as in I-POMDPs \citep{gd05}), `asking' might be modeled. But this would play into the curse of dimensionality; a bad thing.

Clearly, the planning algorithm presented here is highly intractable. Approximate methods for solving large POMDPs could also be looked at to make RepNets practical \citep{dg09,ioz14}.

%Some further questions:
%How would an agent in a RepNet fare if it maximized only reputation, and did not actively pursue positive impact? Would positive impact come automatically by pursuing good reputation? Could positive impact be achieved at the risk of bad reputation through deceit in a RepNet?

%%%%%%%%%%%%%%%%%%%%%%%%%%%%%%%%%%%%%%%%%%%%%%%%%%%%%%%%%%%%%%%%%%%%%%%%%%%%%%%%%%%%%%%%%%%%%%%%%%

\bibliographystyle{apalike}
\bibliography{references}  % put name of your .bib file here

\end{document}